\documentclass[letterpaper, 10 pt, conference]{ieeeconf}  

\usepackage{url}
\usepackage{graphicx}
\usepackage{subcaption}
\usepackage{array} 
\usepackage{algorithm}
\usepackage{algpseudocode}
\usepackage{amsmath} 
\usepackage{amssymb} 
\usepackage{siunitx}
\usepackage{bm}
\usepackage{cite}
\usepackage{hyperref}

\usepackage{caption}
\usepackage{subcaption}
\usepackage[dvipsnames]{xcolor}
\definecolor{green(pigment)}{rgb}{0., 0.75, 0.2}
\usepackage{soul}

\IEEEoverridecommandlockouts                              

\title{\LARGE \bf
Dynamic Object Catching with Quadruped Robot Front Legs
}
\author{André Schakkal, Guillaume Bellegarda, Auke Ijspeert}

\author{André Schakkal, Guillaume Bellegarda, Auke Ijspeert
\thanks{
This research is supported by the Swiss National Science Foundation (SNSF) as part of project No.197237. 
The authors are with the BioRobotics Laboratory, Ecole Polytechnique Federale de Lausanne (EPFL).
{\tt \{firstname.lastname\}@epfl.ch}}}

\begin{document}
\bstctlcite{MyBSTcontrol}

\maketitle
\thispagestyle{empty}
\pagestyle{empty}

\begin{abstract}
This paper presents a framework for dynamic object catching using a quadruped robot's front legs while it stands on its rear legs. The system integrates computer vision, trajectory prediction, and leg control to enable the quadruped to visually detect, track, and successfully catch a thrown object using an onboard camera. Leveraging a fine-tuned YOLOv8 model for object detection and a regression-based trajectory prediction module, the quadruped adapts its front leg positions iteratively to anticipate and intercept the object. The catching maneuver involves identifying the optimal catching position, controlling the front legs with Cartesian PD control, and closing the legs together at the right moment. We propose and validate three different methods for selecting the optimal catching position: 1) intersecting the predicted trajectory with a vertical plane, 2) selecting the point on the predicted trajectory with the minimal distance to the center of the robot’s legs in their nominal position, and 3) selecting the point on the predicted trajectory with the highest likelihood on a Gaussian Mixture Model (GMM) modelling the robot's reachable space. Experimental results demonstrate robust catching capabilities across various scenarios, with the GMM method achieving the best performance, leading to an 80\% catching success rate.
A video demonstration of the system in action can be found at 
\href{https://youtu.be/sm7RdxRfIYg}{https://youtu.be/sm7RdxRfIYg}.
\end{abstract}

\section{Introduction}

Quadruped robots are showing impressive abilities to traverse challenging terrains~\cite{miki2022learning,bellegarda2022cpgrl,bellegarda2024visual}, run at high speeds~\cite{ji2022concurrent,bellegarda2021robust}, and locomote over dynamic parkour obstacles~\cite{shafiee2024viability,cheng2023parkour,zhuang2023robot,shafiee2023puppeteer,shafiee2024manyquadrupeds}. Recently, to allow quadrupeds to accomplish everyday tasks, there are an increasing number of examples of mounting a manipulator arm on top of a quadruped to perform both locomotion and manipulation (loco-manipulation)~\cite{fu2022deep,ferrolho2023roloma,sleiman2023versatile,arcari2023bayesian}. 

This results in five (or more) ``arms'' to perform manipulation and locomotion. On the other hand, bipeds use two feet for locomotion, and typically have two arms for world interactions such as opening doors and moving boxes, also recently emulated with a wheel-legged quadruped balancing on its rear wheeled-legs~\cite{schwarke2023curiosity}. 
Some recent works investigate the possibility of using existing degrees of freedom on typical quadruped robots to perform tasks such as pressing buttons and opening doors~\cite{cheng2023legmanip}, or moving boxes by pushing them from different lateral directions using the body~\cite{rigo2023contact,sombolestan2023hierarchical,jeon2023learning}.
Another approach for interacting with moving or thrown objects is mounting a net on the robot base, either positioned horizontally and leveraging event cameras to detect and catch an object thrown with high speed~\cite{Forrai_23_ICRA}, or mounted vertically to predict the landing location of an object and then moving the robot to make the catch~\cite{you2023run}. Other works show manipulation with multiple legs is possible by resting the quadruped base on the ground, and  using the feet in the air to re-orient a ball~\cite{shi2021circus}, or resting parts of the robot base on the ground, and using the limbs to push or pick up boxes~\cite{wolfslag2020optimisation}. 
\begin{figure}[t]
    \centering
    \includegraphics[width=\linewidth]{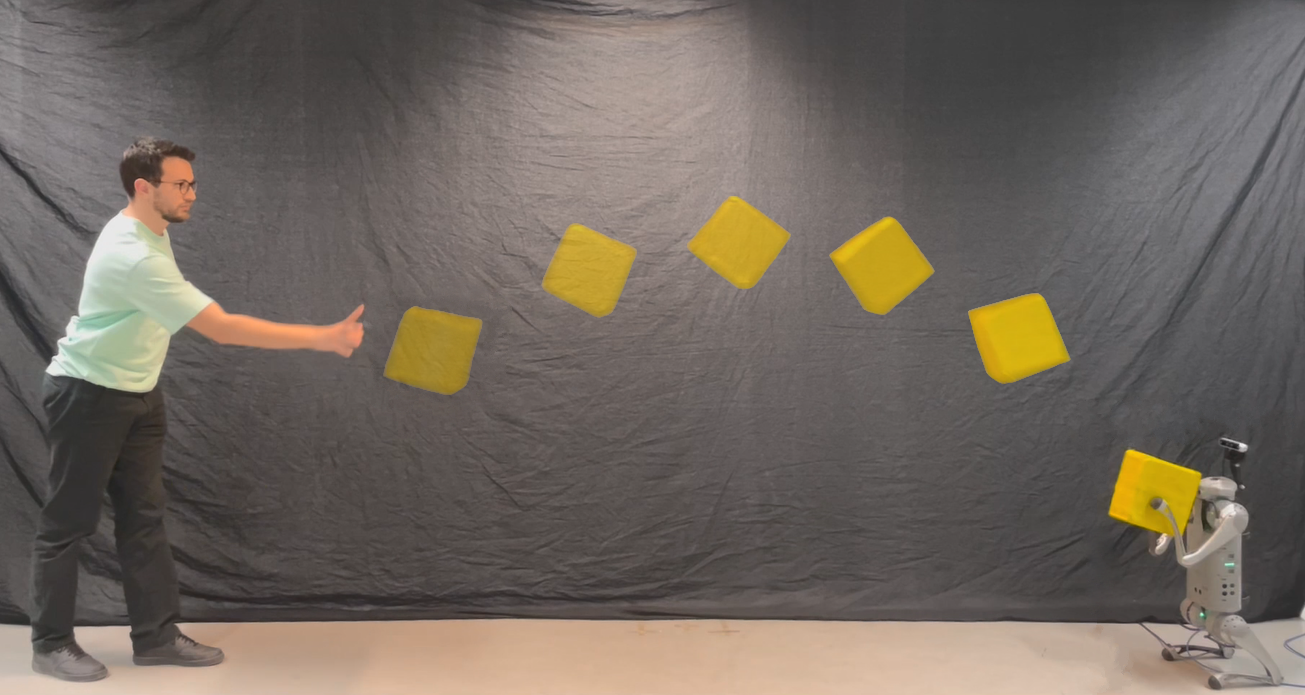}
    \caption{The Unitree Go1 quadruped robot elevated on its rear legs, detecting a thrown object, predicting a suitable catching position, and successfully catching it with its front legs.
    }
    \vspace{-1.9em}
    \label{fig:intro}
\end{figure}

In contrast to previous works, in this paper we consider the task of using a quadruped's existing degrees of freedom to catch a thrown object, as shown in Figure~\ref{fig:intro}. This necessitates reconfiguring the robot from its nominal standing position on four legs, to standing on its rear legs. Once elevated, this new configuration allows for manipulation capabilities by using its two front limbs together to interact with the world, for example to catch a thrown object. 
A camera mounted onboard the robot is used to detect the object, and multiple consecutive images are used to predict a suitable catching position. Upon predicting and selecting this catching position and the time for the object to reach this position, the robot can move its limbs to catch the object.

\section{Related work}\label{related}

Dynamic object catching stands as a rich area of research interest to demonstrate fast prediction and control capabilities, rooted in early pioneering works \cite{hong_experiments_1997},\cite{riley_robot_2002}. Over the years, researchers have explored diverse methodologies and systems, including the use of humanoids \cite{riley_robot_2002,kim_catching_2014,kober_playing_2012}, robotic arms employing various end-effectors such as baskets \cite{dong_catch_2020,frese_off--shelf_2001,gold_catching_2022,huang_offset-free_2021}, grippers \cite{hong_experiments_1997,carneiro_robot_2021}, and robotic hands \cite{salehian_dynamical_2016,lampariello_trajectory_2011,bauml_kinematically_2010,huang_dynamic_2023}. 
Quadrupeds have also been employed for catching objects \cite{Forrai_23_ICRA},\cite{you2023run}. It is noteworthy that the existing approaches involving quadrupeds significantly differ from the methodology proposed in this paper. Specifically, in \cite{Forrai_23_ICRA} and \cite{you2023run}, quadrupeds operate on all fours, catching objects with nets/baskets mounted on their backs. To the best of our knowledge, no prior work has demonstrated an object-catching application with a standing quadruped utilizing its front legs.

Dynamic object catching involves several essential subtasks: object detection and tracking, trajectory prediction, and the catching control maneuver. In the following, we review previous existing works in each of these areas. 

\subsection{Object Detection and Tracking}
Object detection serves as a crucial component in dynamic object catching, enabling the system to determine the location of the thrown object for trajectory prediction and subsequent catching maneuvers. Object tracking methodologies can be broadly classified into two main categories: those requiring depth information in conjunction with the detection module for accurate 3D localization, and those operating without additional depth information.

Among the methods not requiring extra depth information, motion capture systems have been employed, as seen in works utilizing the Vicon motion capture system \cite{dong_catch_2020,carneiro_robot_2021}, and the Optitrack vision system \cite{salehian_dynamical_2016}. While these systems offer high accuracy, they necessitate markers on the thrown object, a complex setup with multiple cameras, can be relatively expensive, and lack mobility and generalizability. Such methods could however serve as benchmarks to assess the performance of the other modules of the system.

Conversely, depth-dependent methods, requiring additional depth information, are generally more cost-effective and compact. Techniques such as simple color detection, coupled with depth data obtained through stereo vision \cite{riley_robot_2002},\cite{kober_playing_2012},\cite{huang_dynamic_2023},\cite{gomez-gonzalez_real_2020}, fall into this category. A drawback of this approach lies in its object dependency, requiring the system to be informed of the object's color. Moreover, the absence of the object's color in the background is essential to avoid interference with the detection module.
Alternative techniques involve various image processing techniques, such as template matching \cite{kao_ball-catching_2021}.  Additionally, deep learning algorithms such as YOLO (You Only Look Once) \cite{redmon_you_2016} have been employed for object detection \cite{huang_creating_2023}. These methods were complemented by depth estimation via stereo vision for correct 3D localization.
Finally, an interesting variation involves object detection using an event camera, where  the object's depth is subsequently estimated based on perceived width and the camera's focal length, eliminating the need for stereo vision \cite{Forrai_23_ICRA}.

\subsection{Trajectory Prediction}
Predicting the trajectory of a thrown object is crucial for enabling a robotic system to catch it successfully. Various methods have been employed to achieve accurate trajectory prediction.

Traditionally, trajectory prediction has been accomplished through regression-based methods. Recursive Ordinary Least Squares (OLS) regression is employed in some studies \cite{Forrai_23_ICRA}, while others enhance this approach by incorporating a regularization term to account for gravity \cite{hong_experiments_1997},\cite{riley_robot_2002}. Gaussian Process Regression has also been utilized to model and predict object trajectories \cite{lampariello_trajectory_2011}. Additionally, simple approaches involving Kalman filters, leveraging known ballistic models, are used for trajectory predictions in various scenarios \cite{you2023run},\cite{kober_playing_2012},\cite{dong_catch_2020},\cite{huang_offset-free_2021},\cite{kao_ball-catching_2021}.

In scenarios demanding prediction of more complex, non-linear trajectories, advanced methods come into play. Deep learning techniques were used, particularly employing bi-directional Long Short-Term Memory (LSTM) networks \cite{zhao_applying_2018}, and conditional generative models \cite{gomez-gonzalez_real_2020}. Another avenue explores Dynamical System estimation 
to model the motion of the object \cite{kim_catching_2014},\cite{salehian_dynamical_2016},\cite{kim_estimating_2012}.

\subsection{Catching Maneuver}
Executing a successful catching maneuver involves two primary capabilities. Firstly, determining the optimal catching position is crucial. This is often accomplished by using the predicted trajectory of the object. A straightforward approach involves identifying the intersection point between the predicted trajectory and a predefined plane where the robot intends to intercept the object \cite{Forrai_23_ICRA},\cite{you2023run},\cite{kober_playing_2012},\cite{kao_ball-catching_2021}. Another effective strategy is to model the reachable space of the robot and select a point on the predicted trajectory within this space. Various models for the reachable space, including geometric shapes, can be used. The point on the predicted trajectory can then be selected by choosing the one closest to the robot's initial position \cite{frese_off--shelf_2001}, the robot's base \cite{hong_experiments_1997}, or its end effector \cite{carneiro_robot_2021}. Alternatively, Gaussian Mixture Models (GMMs) offer a probabilistic approach, allowing the selection of the catching point based on the highest likelihood within the constraints of the optimization problem \cite{kim_catching_2014}. The catching position can also be determined by solving a nonlinear optimization problem which satisfies both robot constraints and object trajectory, and minimizing an energy-based~\cite{lampariello_trajectory_2011} or acceleration-based~\cite{bauml_kinematically_2010} objective function.

The second critical capability is the motion required to reach and catch the object. Catching maneuvers differ significantly based on the robotic platform employed. Although there have been instances of quadrupeds catching objects, they have involved nets/baskets mounted on the robot's back \cite{Forrai_23_ICRA},\cite{you2023run}. Since our focus is on catching objects specifically with the front legs of a quadruped, the strategies are more aligned with those used in humanoid and robotic arm catching scenarios. Catching motion options range from simple PD control to more advanced methods such as Model Predictive Control (MPC) \cite{gold_catching_2022},\cite{huang_offset-free_2021}, learned controllers \cite{dong_catch_2020},\cite{huang_creating_2023}, and the generation of a second-order Dynamical System to govern the robot's motion \cite{kim_catching_2014},\cite{salehian_dynamical_2016}.

\subsection{Contribution}
We present a novel framework that enables quadruped robots to autonomously catch thrown objects using their front legs.
The key components of our framework include:
\begin{itemize}
    \item \textbf{Quadruped Elevation}: We optimize a standing up behavior for the quadruped to stand on its rear legs, leaving its front legs available for catching.
    \item \textbf{Object Detection}: We fine-tune a YOLOv8 model with a specifically curated dataset to be able to detect different objects, using an onboard camera.
    \item \textbf{Trajectory Prediction}: Using the detected coordinates of the thrown object, we predict its trajectory using gravity-informed ordinary least squares
    \item \textbf{Catching Maneuver}: To perform catching, we select a suitable catching position by comparing three different strategies:
    \begin{itemize}
        \item Intersecting the predicted trajectory with a predefined vertical plane
        \item Selecting the point on the predicted trajectory that has a minimal distance to the center of the robot's legs in their nominal position
        \item Selecting the point on the predicted trajectory that has the highest likelihood on a Gaussian Mixture Model (GMM) modeling the robot's reachable space, which was fit over several catching positions collected by a human directing the robot in passive mode
    \end{itemize}
    Upon selecting the catching position, the robot performs the catch with a  Cartesian PD controller.
\end{itemize}

Using this framework, we observe robust catching capabilities in a variety of scenarios, including an 80\% success rate on a test scenario involving 50 throws from a distance of 2 meters in front of the robot.

The rest of this paper is organized as follows. In Section~\ref{sec:method} we present each of the components of our quadruped catching framework and design choices. In Section~\ref{sec:result} we discuss results and analysis from different catching scenarios, with each of the catching position identification methods. Section~\ref{sec:conclusion} concludes the paper and suggests future directions for further work.

\section{Method}
\label{sec:method}
\begin{figure*}[!t]
    \centering
    \includegraphics[width=\linewidth,trim={1cm 10cm 1cm 10cm},clip]{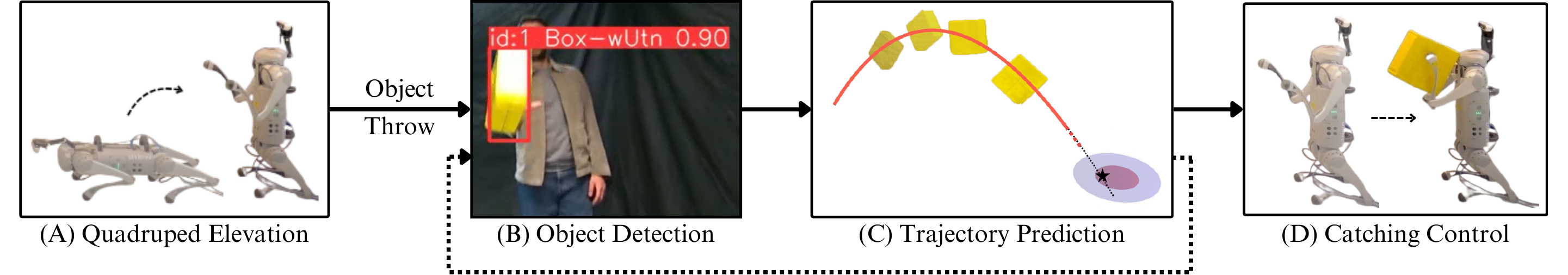} \\
    \vspace{-0.2em}
    \caption{Control diagram for catching objects with a quadruped robot. (A) the quadruped elevates onto its rear legs with an optimized trajectory, leaving its front legs available to catch an object. The user subsequently throws an object. (B) and (C): an onboard camera is used to detect the object and extract its location, the system iteratively predicts a suitable catching location through successive frames while the object is in the air. (D) as the object reaches the predicted catching location, the quadruped controls its front legs to catch the object. 
    }
    \vspace{-1.5em}
    \label{fig:control_diagram}
\end{figure*}

In this section we describe our framework for catching objects with a quadruped robot's front legs. A high-level control diagram is illustrated in Figure~\ref{fig:control_diagram}, showing the four important blocks of our pipeline: (A) elevating the quadruped onto its rear legs, and subsequently throwing the object, (B) detecting and extracting the object location from the frame, (C) using the object locations from successive frames to predict the object's trajectory, and (D) controlling the front legs to catch the object at the predicted catching position.

\subsection{Quadruped Elevation}
In this section we briefly describe the trajectory optimization framework to generate the quadruped elevation motion, shown in Figure~\ref{fig:control_diagram}-(A). It has been adapted from the trajectory optimization framework used in our prior work for generating quadruped jumping motions~\cite{bellegarda_robust_2023,nguyen_optimized_2019}. The robot starts from a standing position with all four feet on the ground, and should end standing on its rear feet, statically balancing with the rear knees in contact with the ground as well. This is accomplished with a time-based two part contact dynamics component: all legs in contact, and then only the rear legs in contact. The  discrete time optimization can be formulated as follows: 
\begin{equation}
\begin{aligned}
\min_{\bm{x}_k,\bm{u}_k;  \ k=1...N }  \quad & J\bigl(\bm{x}_N\bigr) + h\sum_{k=1}^{N} w\bigl(\bm{x}_k,\bm{u}_k\bigr)\\
\textrm{s.t.} \quad & d\bigl(\bm{x}_k,\bm{u}_k,\bm{x}_{k+1}) = 0, \ k=1...N-1\\
  &\phi\bigl(\bm{x}_k,\bm{u}_k\bigr) = 0, \ k=1...N    \\
  &\psi\bigl(\bm{x}_k,\bm{u}_k\bigr) \leq 0, \ k=1...N    \\
\end{aligned}
\end{equation}
\noindent where $\bm{x}_k=\left[p_{x,k};~p_{z,k};~\theta_{k};~\bm{q}_{k}\right]$ is the full state of the system at sample $k$ along the trajectory, $\bm{u}_k$ is the corresponding control input, $J$ and $w$ are final and additive costs to end upright at a particular height while minimizing energy, $h$ is the time between sample points $k$ and $k+1$, and $N$ is the total number of samples along the trajectory. The constraints are specified as follows:
\begin{itemize}
    \item The function $d(\cdot)$ captures the full-body dynamic constraints, which is discretized from
    \begin{equation}\label{eq:full_dynamics_2D}
    \begin{bmatrix}\bm{M} & -\bm{J}_c^T \\ -\bm{J}_c^T & \mathbf{0} \end{bmatrix} \begin{bmatrix}
    \bm{\ddot{x}} \\ \bm{f}_c
    \end{bmatrix}= \begin{bmatrix}-\bm{C}\bm{\dot x}-\bm{g} + \bm{S}\bm{\tau}+\bm{S}_{f}\bm{\tau}_{f} \\ \bm{\dot J}_{c}(\bm{x})\bm{\dot x}\end{bmatrix}, \nonumber
    \end{equation}
    where $\bm{M}$ is the mass matrix, $\bm{C}$ represents the Coriolis  and centrifugal terms, $\bm{g}$ denotes the gravity vector, $\bm{J}_c$ is the spatial Jacobian expressed at the foot contact, $\bm{S}$ and $\bm{S}_{f}$ are distribution matrices of actuator torques $\bm{\tau}$ and joint friction torques $\bm{\tau}_{f}$, $\bm{f}_c$ is the spatial force at the foot contact. The dimensions of $\bm{J}_c$ and $\bm{f}_c$ depend on the contact phases.
    
    \item The function $\phi(\cdot)$ represents equality constraints on initial/final joint and body configurations. 
    
    \item The function $\psi(\cdot)$ captures inequality constraints including joint angle/velocity/torque limits, friction cone limits, and minimum ground reaction forces. 
\end{itemize}
The optimization produces desired joint angles $(\bm{q}_d)$, joint velocities $(\bm{\dot{q}}_d)$ and feed-forward joint torques $(\bm{\tau}_d)$ at a sampling time of 10 \emph{ms}, which are then linearly interpolated to 1 \emph{ms}. These can be tracked to successfully elevate the quadruped with the following joint PD controller running at 1 \emph{kHz} as:
\begin{align}
    \bm{\tau}_{\mathrm{ff}} = \bm{K}_{p,joint} (\bm{q}_d -\bm{q}) + \bm{K}_{d,joint}  (\bm{\dot{q}}_d -\bm{\dot{q}}) + \bm{\tau}_d 
    \label{eqn:jumping_jointPD}
\end{align}
where $\bm{K}_{p,joint}$ and $\bm{K}_{d,joint}$ are proportional and derivative gains in the joint coordinates.

\subsection{Object Detection}\label{detection}
After successfully elevating the quadruped on its rear legs, the subsequent step involves detecting and tracking the thrown object (Figure~\ref{fig:control_diagram}-(B)). For this purpose, an Intel RealSense™ Depth Camera D455 is mounted on the robot, capturing RGB and depth images at 30 frames per second. 

\subsubsection{YOLOv8 Object Detection}\label{yolo_object_detection}

To facilitate object detection and tracking, we use YOLOv8 \cite{redmon_you_2016}. YOLOv8 offers high generalizability in object detection by handling diverse objects without constraints on color or shape, and performs onboard detection without requiring any modifications or attachments to the objects themselves.
We fine-tuned the small pretrained model \texttt{YOLOv8s} from the \texttt{ultralytics} library\footnote{\url{https://github.com/ultralytics/ultralytics}} using a dataset of 389 images. These images were extracted from recorded videos capturing users throwing objects towards the quadruped, with intentional variations in the background to help with generalization. The dataset was annotated using the \texttt{Roboflow} software \cite{dwyer_roboflow_2022}. The recorded videos were processed to generate individual images. The images were then annotated, where bounding boxes were defined around the thrown object, and the object was classified as a ``Box''. Various augmentations were applied to the annotated images, including a horizontal flip, a 15° rotation, and Gaussian blur, all targeted at the bounding boxes. This process generated new data to create a more robust and versatile model. Using this augmented dataset, the model underwent fine-tuning with an AdamW optimizer \cite{loshchilov_decoupled_2019}. The fine-tuning process involved a learning rate of 0.002 and a momentum of 0.9, lasting for 100 epochs. With only one class to detect, this configuration achieved a high mean average precision of 0.992 when
the IoU threshold is set to 0.5 (mAP@50) on the validation set. This allowed us to achieve good object detection with tight bounding boxes around objects of different colors and dimensions.

\subsubsection{Coordinate Transformation}\label{false}

During inference, the fine-tuned model takes as input an RGB image and outputs coordinates of bounding boxes over the detected objects, their classes, and their corresponding confidence values. The coordinates of the bounding box are two dimensional coordinates in pixels in the image frame. The pixel coordinates of the center of the bounding box $x_{p}$ and $y_{p}$ are then transformed into three dimensional coordinates in the robot frame $(x,y,z)$ (the coordinate system is shown in Figure~\ref{fig:catch_pos}) using the depth of the object returned from the RealSense camera $x_{depth}$.
Leveraging the intrinsic parameters of the RealSense camera—specifically $f_x$ and $f_y$ representing the focal length along the image plane's axes, and $pp_x$ and $pp_y$ indicating the pixel coordinates of the principal point in the image plane—we calculate $(x,y,z)$ with the following formulas:
\begin{equation}
x = x_{depth} \cos \theta -  x_{depth} \frac{\left(y_{p} - pp_y\right)}{f_y} \sin \theta 
\end{equation}
\begin{equation}
y = x_{depth} \frac{\left(x_{p} - pp_x\right)}{f_x}
\end{equation}
\begin{equation}
z = - \left( x_{depth} \sin \theta +  x_{depth} \frac{\left(y_{p} - pp_y\right)}{f_y} \cos \theta \right)
\end{equation}
\noindent with $\theta$ being the angle the camera makes with the horizontal.

\subsection{Trajectory Prediction}

With the sequence of object coordinates in the robot frame available, the next step is trajectory prediction (Figure~\ref{fig:control_diagram}-(C)). Accurately predicting the object's path is crucial for determining a suitable catching location. 

\subsubsection{Approach}

We choose to employ gravity-informed ordinary least squares for trajectory prediction due to its effectiveness in modeling ballistic trajectories and computational efficiency, making it well-suited for real-time prediction. Assuming that, upon leaving the user's hand, the object experiences only free-fall motion with an initial velocity, and neglecting air resistance, we know that the object's trajectory will consistently follow a three-dimensional parabola with:
\begin{align}
x\left(t\right) &= v_{0x} t + x_0 \\
y\left(t\right) &= v_{0y} t + y_0 \\
z\left(t\right) &= -\frac{1}{2}g t^2 + v_{0z} t + z_0
\end{align}
where  $(x_0,y_0,z_0)$ is the initial position of the object at the start of the trajectory, $(v_{0x},v_{0y},v_{0z})$ is the initial velocity, and $g$ is gravity. 

\subsubsection{Parameter Estimation}

Since measurements inevitably contain some degree of noise, exact parameters for the equations above cannot be determined. Therefore, we solve the following system of equations:
\begin{align}
x_{i}\left(t_i\right) &= a_x t_i + b_x, \quad i = 1,..., n, \\
y_{i}\left(t_i\right) &= a_y t_i + b_y, \quad i = 1,..., n, \\
z_{i}\left(t_i\right) &= a_z t_i^2 + b_z t_i + c_z, \quad i = 1,..., n,
\label{yeq}
\end{align}
where $a$, $b$, $c$ represent the coefficients to be estimated, and $n$ is the number of measurements. We have that $n \geq 3$ since for Equation \ref{yeq}, we need at least three measurements to have a solution for the parameters of the equation.

For $x$ and $y$, we find their corresponding parameters by solving the following normal system of equations:
\begin{equation}
\begin{bmatrix}
\sum_{i=1}^{n} 1 & \sum_{i=1}^{n} t_i\\
\sum_{i=1}^{n} t_i & \sum_{i=1}^{n} t_i^2\\
\end{bmatrix}
\begin{bmatrix}
b_x \\
a_x \\
\end{bmatrix}
=
\begin{bmatrix}
\sum_{i=1}^{n} x_{i} \\
\sum_{i=1}^{n} x_{i} t_i \\
\end{bmatrix}
\end{equation}
\begin{equation}
\begin{bmatrix}
\sum_{i=1}^{n} 1 & \sum_{i=1}^{n} t_i\\
\sum_{i=1}^{n} t_i & \sum_{i=1}^{n} t_i^2\\
\end{bmatrix}
\begin{bmatrix}
b_y \\
a_y \\
\end{bmatrix}
=
\begin{bmatrix}
\sum_{i=1}^{n} y_{i} \\
\sum_{i=1}^{n} y_{i} t_i \\
\end{bmatrix}
\end{equation}

On other hand, since $z$ follows a parabolic trajectory, we determine its corresponding parameters by solving the following system of normal equations, introducing $\lambda=1$ to account for gravity:
\vspace{-1.2em}

\begin{multline}
\begin{bmatrix}
\sum_{i=1}^{n} 1 & \sum_{i=1}^{n} t_i & \sum_{i=1}^{n} t_i^2\\
\sum_{i=1}^{n} t_i & \sum_{i=1}^{n} t_i^2 & \sum_{i=1}^{n} t_i^3\\
\sum_{i=1}^{n} t_i^2 & \sum_{i=1}^{n} t_i^3 & \sum_{i=1}^{n} t_i^4 + \lambda\\
\end{bmatrix}
\begin{bmatrix}
c_z \\
b_z \\
a_z \\
\end{bmatrix} \\
=
\begin{bmatrix}
\sum_{i=1}^{n} z_{i} \\
\sum_{i=1}^{n} z_{i} t_i \\
\sum_{i=1}^{n} z_{i} t_i^2 - \lambda \frac{1}{2}g \\
\end{bmatrix}
\label{eq:your_equation}
\end{multline}

\subsubsection{Iterative Refinement}
Solving this system of equations provides a time-dependent model $\bm{x}_{pred}\left(t\right) = \left(x_{pred}\left(t\right),y_{pred}\left(t\right),z_{pred}\left(t\right) \right) $. Consequently, we can predict the position of the object for a specific timestep $t$. Similarly, we can determine the timestep $t$ at which the object will reach a specific $(x,y,z)$.

\begin{figure}[t]
    \centering
    \includegraphics[width = \columnwidth]{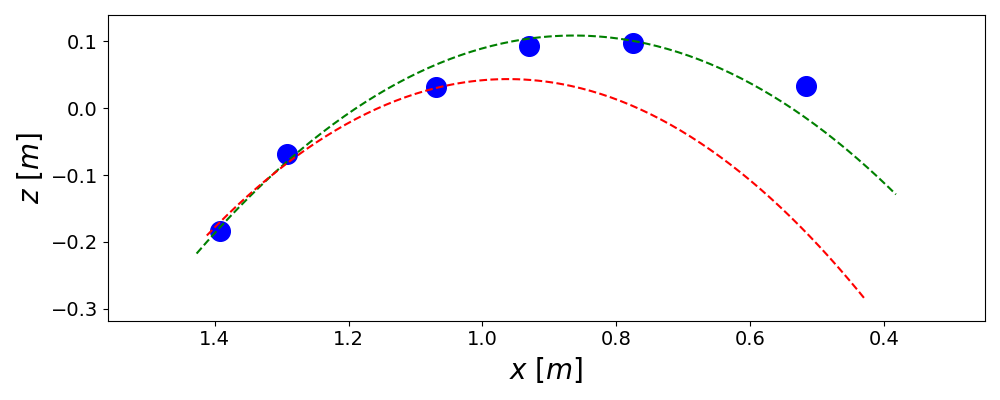}\\
    \vspace{-0.5em}
    \caption{Example trajectory prediction. The discrete blue points represent observed object positions, the red curve shows the predicted trajectory using the first 3 observed positions, and the green curve shows the predicted trajectory using all 6 observed positions.}  
    \label{fig:traj_pred}
    \vspace{-1.2em}
\end{figure}

For each new frame captured by the camera providing new $(x,y,z)$ measurements, we re-solve the system of equations to refine the estimation of the regression parameters. The iterative nature of this process ensures that with each new measurement, we enhance the accuracy of the regression parameters by incorporating more data points into the regression. Such improvement in performance is shown in Figure \ref{fig:traj_pred}, where the addition of more measurements to the prediction module results in predictions closer to the ground truth.

\subsubsection{Initiating Trajectory Prediction}
An important point for correctly starting the trajectory prediction is that the object has been thrown and left the user's hand. This distinction prevents the coordinates of the object, determined when it is in the user's hands, from being considered in the trajectory prediction module, which would lead to incorrect predictions. To address this, the system considers the detected object coordinates for prediction only when the absolute difference between the coordinates of the current and previous detection exceeds a predefined threshold. 

\subsection{Catching Maneuver}
\label{sec:catching_maneuver}
Having achieved the ability to detect, track, and predict the trajectory of the thrown object, the final submodule essential for a successful catch is the catching maneuver itself (Figure~\ref{fig:control_diagram}-(D)). 

\subsubsection{Catching Position Identification}
\begin{figure}[t]
    \centering
    \begin{subfigure}{0.32\columnwidth} 
        \centering
        \includegraphics[width=\linewidth]{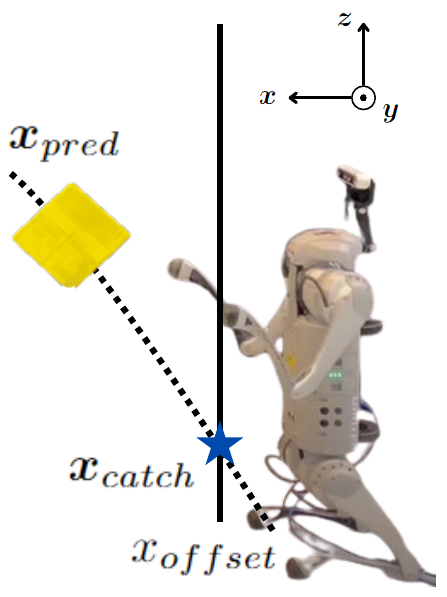}
        \caption{Plane Intersection}
        \label{subfig:method_a}
    \end{subfigure}
    \hfill
    \begin{subfigure}{0.32\columnwidth} 
        \centering
        \includegraphics[width=\linewidth]{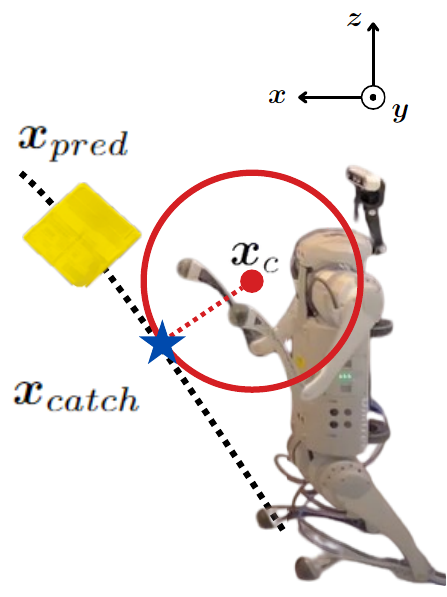}
        \caption{Min Distance to Center}
        \label{subfig:method_b}
    \end{subfigure}
    \hfill
    \begin{subfigure}{0.32\columnwidth} 
        \centering
        \includegraphics[width=\linewidth]{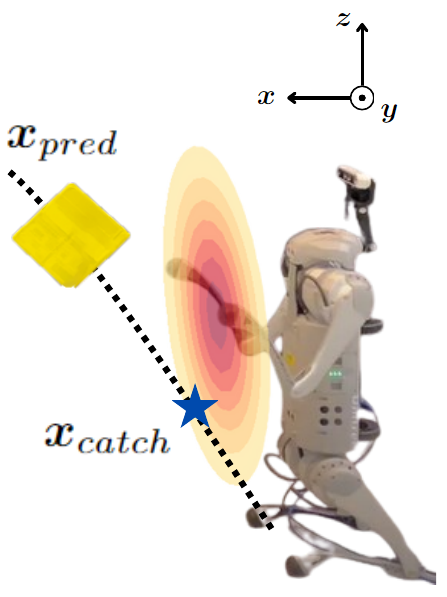}
        \caption{GMM}
        \label{subfig:method_c}
    \end{subfigure}
    \caption{Illustrations of the different catching position identification methods.}
    \label{fig:catch_pos}
    \vspace{-1.9em}
\end{figure}

There are a potentially infinite number of possible positions where the quadruped can catch the thrown object. In this paper, we consider three different methods for determining a suitable catching position $\bm{x}_{catch} = \left(x_{catch}, y_{catch}, z_{catch}\right)$, each illustrated in Figure~\ref{fig:catch_pos}.

\begin{figure}[t]
    \centering
    \begin{subfigure}{0.492\columnwidth}
        \centering
        \includegraphics[width=\linewidth]{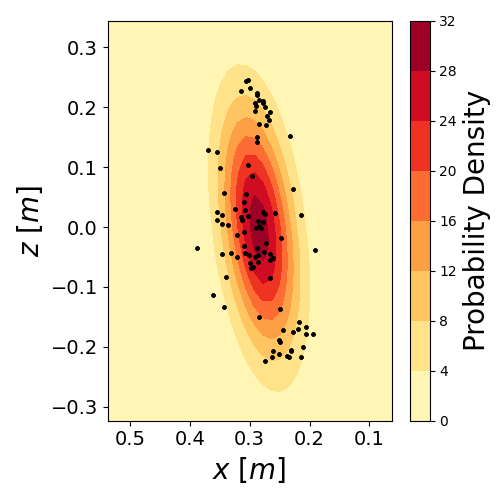}\\
        \vspace{-0.5em}
        \caption{Side View GMM}
        \label{subfig:a}
    \end{subfigure}
    \hfill
    \begin{subfigure}{0.492\columnwidth}
        \centering
        \includegraphics[width=\linewidth]{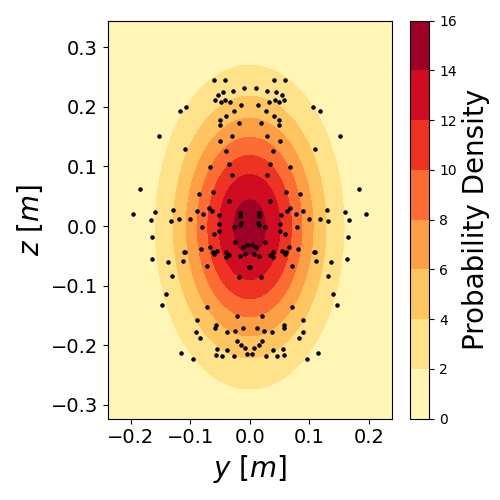}\\
        \vspace{-0.5em}
        \caption{Front View GMM}
        \label{subfig:b}
    \end{subfigure}
    \caption{GMM fitted over 100 catching demonstrations. The GMM models the quadruped's reachable space and provides likelihood information for catching positions.}
    \label{fig:GMM}
    \vspace{-1.7em}
\end{figure}

\begin{itemize}
    \item \textbf{Plane Intersection}:  
    The catching position $\bm{x}_{catch}$ and the time to reach this position $t_{catch}$ are determined by intersecting the predicted trajectory with a vertical plane $x_{offset} = 25cm $ in front of the quadruped, as illustrated in Figure \ref{subfig:method_a}. Specifically, we find the timestep $t_{catch}$ that solves:
\begin{equation}
    x_{pred}\left(t_{catch}\right) = x_{offset} \ \Leftrightarrow \ t_{catch} = \frac{x_{offset} - b_x}{a_x} 
\end{equation}

Subsequently, we determine the catching position $\bm{x}_{catch}$ by substituting this timestep $t_{catch}$ into the trajectory equations:
\begin{equation}
\bm{x}_{catch} = \bm{x}_{pred}\left(t_{catch}\right)
\end{equation}
    \item \textbf{Minimum Distance to Center}: The catching position $\bm{x}_{catch}$ is determined by choosing the point on the predicted trajectory that has the smallest Euclidean distance to the center between the front feet of the quadruped in its nominal configuration $\bm{x}_c$. The catching position is thus the closest point to the original center between the front feet, which is on the predicted trajectory, as shown in Figure \ref{subfig:method_b}. Finding the catching position $\bm{x}_{catch}$ and the time to reach this position $t_{catch}$ is equivalent to solving the following optimization problem:
    \begin{equation}
    \begin{aligned}
    \bm{x}_{catch},t_{catch} = \min_{\bm{x},t}  \quad & \lVert\bm{x}_c - \bm{x} \rVert_2 \\
    \textrm{s.t.} \quad & \bm{x}_{pred}\left(t\right) = \bm{x} \\
    \end{aligned}
    \end{equation}
    
    \item \textbf{Gaussian Mixture Model (GMM)}: The third approach consists of using Gaussian Mixture Models (GMMs). Inspired by \cite{kim_catching_2014}, we performed 100 catching demonstrations by throwing the object towards the quadruped, and a human manually directed the quadruped to catch the object with its front feet. The final positions where the object was caught were recorded.  We then fit a GMM on these points to have a density distribution of these points. The parameters of the GMM $\{\pi_k, \bm{\mu}_k, \bm{\Sigma}_k\}_{k=1:K}$, where $K$ is the number of Gaussians, were determined through expectation maximization. Using the Bayesian information criterion (BIC), the optimal number of Gaussians was determined to be $K=1$, therefore $\pi_1=\pi=1$. The probability density of a catching position $\bm{x}$ is shown in Figure \ref{fig:GMM} and is given by
\begin{equation}
    \mathcal{P}\left(\bm{x}|\bm{\mu}, \bm{\Sigma}\right) = \mathcal{N}\left(\bm{x}|\bm{\mu}, \bm{\Sigma}\right).
\end{equation}
The fitted GMM not only implicitly covers the reachable workspace of the quadruped's front legs and gives suitable catching locations, but it also gives us likelihood information. Regions where the likelihood is high are regions where the human driving the quadruped during the demonstrations caught the object many times. Since the quadruped could possibly catch the object on many positions along the predicted trajectory, this likelihood information leads to choosing a more human-intuitive catching position. Having this probabilistic model, the catching position is chosen to be the point on the predicted trajectory that has the highest likelihood on the probability density of the GMM, as shown in Figure \ref{subfig:method_c}. Therefore, finding the catching position $\bm{x}_{catch}$ and the time to reach this position $t_{catch}$ is equivalent to solving the following optimization problem:
\begin{equation}
\begin{aligned}
\bm{x}_{catch},t_{catch} = \max_{\bm{x},t}  \quad & \mathcal{P}\left(\bm{x}|\bm{\mu}, \bm{\Sigma}\right)\\
\textrm{s.t.} \quad & \bm{x}_{pred}\left(t\right) = \bm{x} \\
\end{aligned}
\end{equation}

\end{itemize}

\noindent 

For each method explained above, $\bm{x}_{catch}$ and $t_{catch}$ are updated for every new camera frame providing new measurements in order to improve the catching position prediction.
Furthermore, using the time to reach the catching position $t_{catch}$ and the current timestep $t$, we can calculate the time remaining for the object to reach the catching position:
\begin{equation}
    t_{remain} = t_{catch} - t
\end{equation}

\subsubsection{Catching Control}

As the prediction is iteratively refined, we use Cartesian PD control to close the legs around the predicted catching position $\bm{x}_{catch}$. These coordinates are mapped to leg frame coordinates $\bm{p}_d = (x_{catch}, y_{catch} \pm y_{opened}, z_{catch})$, where $y_{opened}=0.15m$ ensures that the front legs are opened around the predicted catching position. These values are first clipped to ensure they remain in the robot workspace in case the prediction makes the catch impossible.

The subsequent step is to determine when to close the feet, and this is done using $t_{remain}$. Since the feet of the quadruped are always positioned around the predicted catching position, and the time remaining for the object to reach this position is known, the closing mechanism is activated when $t_{remain}$ drops below a time threshold $t_{thresh}$. Through experimentation, the optimal time threshold was determined to be $t_{thresh} = 0.15s $.

As soon as $t_{remain}$ drops below $t_{thresh}$, the feet coordinates in the robot frame are offset to close around the object with $\bm{p}_d = (x_{catch}, y_{catch} \pm y_{closed}, z_{catch})$, where $y_{closed}=0.01m$ ensures that the front legs close around the object. The resulting torque control for each front leg is written as follows: 
\begin{align}
\bm{\tau} &= \bm{J}(\bm{q})^\top \Bigl[\bm{K}_{p} \left(\bm{p}_{d} - \bm{p} \right) - \bm{K}_{d} \left( \bm{v} \right)  \Bigr] - \bm{K}_{d,joint} (\bm{\dot{q}})
\end{align}
where $\bm{J}(\bm{q})$ is the foot Jacobian at joint configuration $\bm{q}$, $\bm{K}_p$ and $\bm{K}_d$ are diagonal matrices of proportional and derivative gains in Cartesian coordinates to track the desired foot positions $(\bm{p}_d)$ with zero desired foot velocity $(\bm{v})$ in the leg frame. We also add a small joint damping term for stability. We use $\bm{K}_{p}=400\bm{I}_3,\ \bm{K}_{d}=8\bm{I}_3, \  \bm{K}_{d,joint}=\bm{I}_3$.

\subsection{Experimental Setup}
\begin{figure}[t]
    \centering
    \begin{subfigure}{0.24\columnwidth} 
        \centering
        \includegraphics[width=4.15\linewidth,trim={2cm 7cm 2cm 7cm},clip]{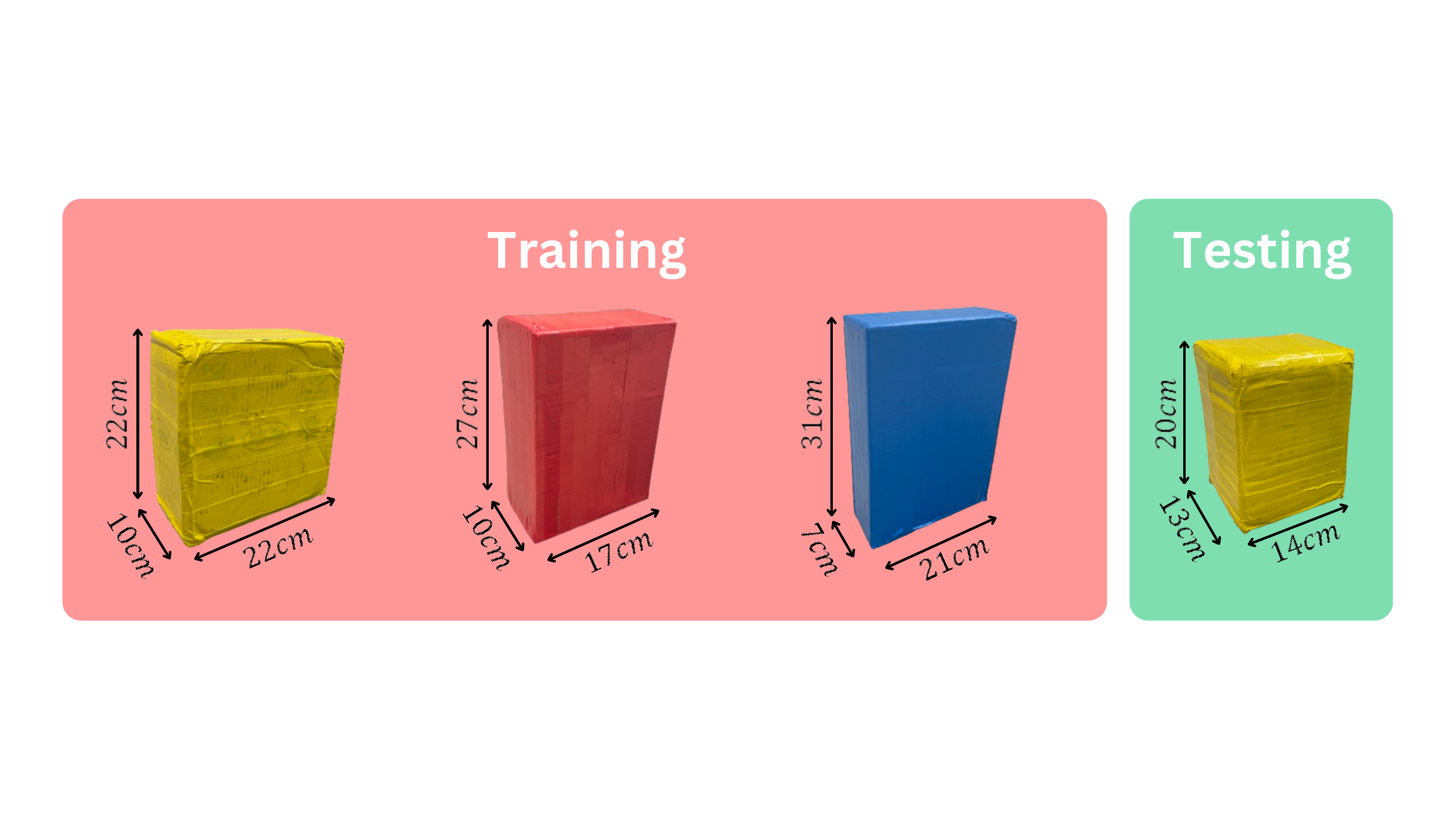}
        \caption{Object A}
        \label{subfig:objA}
    \end{subfigure}
    \hfill
    \begin{subfigure}{0.24\columnwidth} 
        \centering
        \caption{Object B}
        \label{subfig:objB}
    \end{subfigure}
    \hfill
    \begin{subfigure}{0.24\columnwidth} 
        \centering
        \caption{Object C}
        \label{subfig:objC}
    \end{subfigure}
    \hfill
    \begin{subfigure}{0.24\columnwidth} 
        \centering
        \caption{Object D}
        \label{subfig:objD}
    \end{subfigure}
    \caption{Objects used in experiments for both training and testing.}
    \label{fig:obj}
    \vspace{-1.5em}
\end{figure}

We use the Unitree Go1 quadruped~\cite{unitreeGO1}, on which we mount a depth camera for object detection and position tracking (Intel® RealSense™ Depth Camera D455), accelerated on a graphics processing unit (GPU) for fast calculations (Nvidia RTX 1050i GPU). Throughout the experiments, we use several objects (shown in Figure \ref{fig:obj}) with different colors and dimensions, ranging in mass from 150g to 200g. Objects A, B and C were used for training the object detection module and tuning the catching time threshold $t_{thresh}$. Object D, on the other hand, is used in the following section for testing purposes.

\section{Results}
\label{sec:result}
In this section, we discuss results from using our framework to catch various objects thrown towards the robot. Example snapshots of a successful catch are shown in Figure~\ref{fig:intro}. 
The reader is also encouraged to watch the supplementary video, which illustrates the framework's effectiveness through diverse tests with various objects of different sizes and colors.
In particular, with our experiments we seek to answer the following questions:
\begin{itemize}
    \item How robust is the framework to varying throwing styles (i.e. possible catching locations in the robot workspace), and to different thrown objects? 
    \item What are the effects and benefits of the different catching position intersection methods presented in Section~\ref{sec:catching_maneuver}?
\end{itemize}

\subsection{General Performance}
To assess the overall performance of our system, we conducted catching experiments using object D which was not included in the YOLOv8 model's training data during finetuning. Object D was used for testing due to its smaller catching surface, enabling us to evaluate the different catching position identification methods outlined in Section~\ref{sec:catching_maneuver}. For each of the three methods, we threw the object 50 times from a distance of 2 meters aiming at the center of the quadruped's front legs. A successful catch was defined as the quadruped closing its front legs to hold the thrown object without it falling to the ground. 
Table~\ref{table_results} shows the success rates and the average total power consumed for successful catches, across the 50 throws for each of the three methods.

We observe an 8\% difference in performance between the Plane Intersection method and the Minimum Distance to Center method, and a 2\% difference in performance between the Minimum Distance to Center and the GMM method. Clearly, the latter two methods significantly enhance the catching performance of the quadruped compared with the simpler Plane Intersection baseline method. Moreover, comparing the Plane Intersection method with the Minimum Distance to Center and GMM methods shows a 19\% and 31\% decrease in average power consumption, respectively.
In summary, we find that for ``easy-to-catch'' throws aimed at the center, the Minimum Distance to Center and GMM methods outperform the Plane Intersection method. Additionally, these two methods exhibit comparable performance, with the GMM method showing slightly better performance in terms of catching success rate and average power consumption.

\begin{table}
\renewcommand{\arraystretch}{1.3}
\caption{Performance of different catching position identification methods.}
\label{table_results}
\centering
\begin{tabular}{>{\arraybackslash}m{2.7cm}|>{\centering\arraybackslash}m{1.5cm}|>{\centering\arraybackslash}m{1.4cm}|>{\centering\arraybackslash}m{1.3cm}}
\hline
\bfseries & \bfseries Plane Intersection & \bfseries Min Distance to Center & \bfseries GMM \\
\hline\hline
\bfseries Success rate [\%] & 70 & 78 & \bf{80}\\
\hline
\bfseries Mean total power [W] & 21.89 & 17.70 & \bf{15.13}\\
\hline
\end{tabular}
\vspace{-1.7em}
\end{table}

\subsection{Challenging Catching Scenario Case Study}
To assess the system performance under challenging conditions, we evaluated the catching success rate of the different methods with ``harder-to-catch'' throws using object D.
We conducted multiple throws aimed at the limits of the robot's reachable space and observed the success rate of each method. Throws directed toward the rightward, leftward, and upward limits of the workspace yielded similar results to those presented in Table \ref{table_results}. However, throws aimed at the lower limit proved to be more difficult to catch. To evaluate the performance of the methods with these throws, we executed 10 throws aimed lower than 15 cm below the quadruped's shoulders for each method and recorded the success rate. This type of low throw is illustrated in Figure \ref{fig:catch_pos}, where the object is aimed at a low point of the quadruped's reachable space.
Using the Plane Intersection and Min Distance to Center methods, the quadruped failed to catch any low throws. In contrast, employing the GMM method resulted in a 60\% success rate, with the quadruped successfully catching 6 out of the 10 low throws.
The GMM method notably outperforms the other methods with hard-to-catch throws. Its success can be attributed to its probabilistic modeling of the workspace of the front legs and providing insights into more human-intuitive catching locations. Consequently, this method selects the most intuitive catching position on the predicted trajectory, as shown in Figure \ref{subfig:method_c}. Conversely, the failure of the Plane Intersection method to catch low throws can be attributed to the predicted trajectory intersecting the catching plane lower than the front legs' reachable space limits as depicted in Figure \ref{subfig:method_a}. Additionally, the Minimum Distance to Center method, lacking information on catching intuition or the reachable space of the front legs, may select a catching position closest to the nominal center, but in a location that is challenging to catch the object, as shown in Figure \ref{subfig:method_b}.

\section{Conclusion}
\label{sec:conclusion}
In conclusion, this paper presents a comprehensive framework for enabling a quadruped robot to autonomously catch thrown objects using its front legs. The system involves multiple submodules, including quadruped elevation, object detection, trajectory prediction, catching position selection, and the catching control maneuver. 
The YOLOv8-based object detection module, fine-tuned for the specific task, demonstrated effective detection capabilities with tight bounding boxes. 
Trajectory prediction using an ordinary least squares approach showcased iterative refinement of regression parameters, resulting in accurate predictions with increasing numbers of measurements. To select a suitable catching position, we evaluated three different strategies: 1) trajectory intersection with a vertical plane, 2) the closest point on the predicted trajectory to the nominal center of the feet, and 3) the point on the predicted trajectory which has the highest likelihood on a Gaussian Mixture Model modelling the robot's reachable space. We found that the GMM offered the best performance (80\% success rate and lowest power consumption) with easy-to-catch throws, and additionally allowed catches in harder scenarios with boxes thrown at the extreme limits of the robot workspace. 

Limitations include the need for a more controlled experimental setup to fully quantify performance, 
and consideration of the thrown object's orientation on catching performance. 
Future work could focus on improving these limitations, generalization to catch arbitrary objects, and developing more sophisticated control methods (beyond the Cartesian PD closing mechanism, such as MPC) to catch objects thrown at an angle.

\bibliographystyle{IEEEtran}
\bibliography{IEEEabrv,refs1,refs2}

\end{document}